\documentclass{article}

\usepackage{arxiv}

\usepackage[utf8]{inputenc} 
\usepackage[T1]{fontenc}    
\usepackage{booktabs}       
\usepackage{amsfonts}       
\usepackage{nicefrac}       
\usepackage{microtype}      
\usepackage{lipsum}
\usepackage{url}
\usepackage{graphicx}
\usepackage{graphicx}
\usepackage{amsmath}
\usepackage{changepage}
\usepackage{adjustbox}
\usepackage{subfig}
\usepackage{graphicx}
\usepackage{floatrow}
\usepackage{multirow}
\usepackage{graphicx}
\usepackage{color, colortbl}
\usepackage{rotating}
\usepackage{caption}
\usepackage{amsfonts}
\usepackage{adjustbox}

\usepackage{algorithmic}
\usepackage[ruled, noend]{algorithm2e}
\usepackage{amssymb}
\usepackage{tikz}

\definecolor{Gray}{gray}{0.9}

\begin{document}

\title{Evolutionary Time-Use Optimization for Improving Children's Health Outcomes}

\author{
  Yue Xie, Aneta Neumann \\
  The University of Adelaide\\ Adelaide, SA, Australia \\
  \texttt{yue.xie@adelaide.edu.au} \\
   \And
  Ty Stanford \\
  Alliance for Research in Exercise, \\ Nutrition and Activity, \\ Allied Health and Human Performance,\\ University of South Australia \\
  Adelaide, SA, Australia \\
   \AND
   Charlotte Lund Rasmussen \\
  Norwegian University of Science and Technology \\
   Department of Physical Education and Sport Sciences, University of Limerick \\
   \And
   Dorothea Dumuid \\
   Alliance for Research in Exercise, \\ Nutrition and Activity, \\ Allied Health and Human Performance,\\ University of South Australia \\
  Adelaide, SA, Australia \\
   \And
   Frank Neumann \\
   The University of Adelaide\\ Adelaide, SA, Australia \\
}

\maketitle

\begin{abstract}
How someone allocates their time is important to their health and well-being. In this paper, we show how evolutionary algorithms can be used to promote health and well-being by optimizing time usage. Based on data from a large population-based child cohort, we design fitness functions to explain health outcomes and introduce constraints for viable time plans. We then investigate the performance of evolutionary algorithms to optimize time use for four individual health outcomes with hypothetical children with different day structures. As the four health outcomes are competing for time allocations, we study how to optimize multiple health outcomes simultaneously in the form of a multi-objective optimization problem. We optimize one-week time-use plans using evolutionary multi-objective algorithms and point out the trade-offs achievable with respect to different health outcomes.

\end{abstract}

\section{Introduction}

Evolutionary algorithms (EAs) are bio-inspired randomized optimization techniques and have been very successfully applied to various real-world combinatorial optimization problems \cite{DBLP:journals/memetic/HanW21,DBLP:journals/corr/abs-2107-11300,DBLP:conf/cec/LiBMB13}. Evolutionary algorithms use a population of search points in the decision space of a given optimization problem to solve the problem. Moreover, many real-world optimization problems consist of several conflicting objectives that must be optimized simultaneously. No single solution can optimize multiple objectives, instead a set of trade-off optimal solutions is obtained. EAs can approximate multiple optimal solutions in a single run, which make EAs popular in solving multi-objective optimization problems \cite{DBLP:books/daglib/0010311,DBLP:books/daglib/0004745}.

A real-world multi-objective optimization problem is "How should children spend their time (i.e. sleeping, sedentary behaviour and physical activity) to optimize their health, well-being, and cognitive development?" \cite{WOS:000379430600004,WOS:000379430600003}. The importance of this problem has led governing bodies and health authorities such as the World Health Organization (WHO) to provide guidelines for daily durations of sleep, screen time, and physical activity \cite{okely2022collaborative}. Such guidelines for school-aged children (5-12 years) currently recommend 9-11 hours of sleep, no more than 2 hours of sedentary screen time, and at least 1 hour of moderate-to-vigorous physical activity (MVPA) per day \cite{okely2022collaborative}. However, these guidelines are primarily underpinned by systematic reviews collating evidence of how the duration of a single behaviour, such as MVPA, is associated with a single measure of health or wellbeing \cite{okely2022collaborative}. These studies show whether more or less of behaviour is beneficially associated with the outcome \cite{WOS:000379430600003,WOS:000379430600004,okely2022collaborative}, rather than identifying optimal durations, which would be required to support recommendations for daily durations of the behaviour. Almost no studies have attempted to define optimal durations for these activity behaviours for a single health outcome, let alone for multiple health and well-being outcomes. 

To address the lack of evidence for optimal time-use allocations, a recent study \cite{dumuid2021goldilocks} used compositional linear regression \cite{WOS:000452307300013} to model the relationship between how children allocated their daily time to four activities (sleep, sedentary behaviour, light physical activity (LPA) and MVPA) and twelve outcomes spanning physical, mental and cognitive health domains. Compositional data analysis enabled all four activities to be included in a single model whilst ensuring their constant-sum constraint to 24 hours was respected \cite{aitchison1982statistical}. Using published compositional data methods, the raw activity data of minutes per day were expressed as a set of isometric log-ratios \cite{mateu2011principle}. With these compositional regression models, \cite{WOS:000611969200046} estimated values of the outcomes for every possible and feasible combination of sleep, sedentary behaviour, LPA and MVPA duration were calculated. Optimal daily duration of the activities were derived for each of the twelve health outcomes from the average “time-use composition” associated with the best $5\%$ of estimated values for the respective health outcomes. 

It remains unknown how to perform the best multi-objective optimisation of time use for overall health and well-being. The method developed by \cite{WOS:000611969200046} is computationally intensive for four activities requiring almost 4 million iterations of different possible time-use scenarios. This method becomes unfeasible with a large number of daily activities (e.g., activities such as chores, sport, transport, school, sleep, quiet time, social time, screen time, etc.) routinely collected by time-use recalls \cite{WOS:000209772600010}. Additionally, varying constraints to daily time use, which may limit application to the real world, were not considered. 

The research described in this paper extends previous work proposed in \cite{dumuid2021goldilocks} by considering four decision variables: daily time allocation to sleep, sedentary behaviour, LPA and MVPA, and four health objectives for children: body mass index (BMI), cognition, life satisfaction and fitness. Firstly, we formulate the one-day time-use optimization problem as a single-objective problem in continuous space by optimizing one of the four presented health outcomes. Then, we extend the one-day time-use schedule to one week and present multi-objective optimization models for the time-use optimization problem.

EAs are introduced to develop time-use optimization approaches that incorporate daily and weekly time constraint schedules and provide decision-making tools for trading off multiple health outcomes against each other. For single-objective time-use optimization, we evaluate the performance of the differential evolution (DE) algorithm \cite{storn1997differential} with different operators, particle swarm optimization (PSO) \cite{WOS:A1995BF46H00374} and covariance matrix adaptation evolutionary strategy (CMA-ES)\cite{DBLP:journals/ec/HansenO01,DBLP:books/sp/06/Hansen06} to optimize health outcomes in different day structures. For multi-objective time-use optimization, we investigate the performance of the multi-objective evolutionary algorithm based on decomposition (MOEA/D) \cite{DBLP:journals/tec/ZhangL07}, Non-dominated sorting genetic algorithm (NSGA-II) \cite{996017} and Strong Pareto evolutionary algorithm 2 (SPEA2) \cite{zitzler2001spea2}. 

The paper is organized as follows. We introduce the data set used in Section 1.1. Section \ref{sec:model} describes application of our time-use optimization models for different health outcomes, and to different day constraints. The proposed optimization methods are described in Section \ref{sec:algoithm}. The results of the optimization experiments are described in Section \ref{sec:exp}. Conclusions and avenues for future work are presented in Section \ref{sec:con}.

\subsection{Data Description}
\label{subsec:data}

This study uses data from a large population-based child cohort to illustrate the real-world application of a novel time-use optimisation procedure. Data were from the Child Health CheckPoint study \cite{WOS:000477729200002}, a cross-sectional module nested between waves 6 and 7 of the Longitudinal Study of Australian Children (LSAC) \cite{WOS:000213786500001}. Child participants of the LSAC birth cohort (commenced in 2004 with n=5107) that were retained to Wave 6 $(n=3764)$ were invited to take part in Child Health CheckPoint (2015-16) when they were 11-12 years old. Of these, $n=1874$ $(50\%)$ consented to participate via written informed consent from their parent/guardian. Ethical approval for CheckPoint was granted by The Royal Children’s Hospital (Melbourne) Human Research Ethics Committee (HREC33225D) and the Australian Institute of Family Studies Ethics Committee (AIFS14-26).

Participants were fitted with a wrist-worn accelerometer (GENEActive, Activinsights Ltd, UK) by a trained researcher, with instructions to wear the device 24 hours a day for eight days. Following the return of the device, activity data were downloaded and processed following published procedures \cite{dumuid2021goldilocks,WOS:000213786500001} to determine the average daily minutes spent in sleep, sedentary time, LPA and MVPA. 

BMI was derived from the child participant’s measured height (Invicta 10955 stadiometer) and weight (2-limb Tanita BC-351 or 4-limb InBody 230). BMI was calculated as weight (kg)/height $(m)^2$ and expressed as age- and sex-specific z-scores \cite{WOS:000250059700008}. The cognition score was derived from the NIH Picture Vocab test, which asks the child to select on an iPad a picture that best represents the meaning of words they hear through headphones \cite{WOS:000315995100012}. A higher score indicates better receptive vocabulary, which represents cognition. Life satisfaction was obtained from the 5-item Brief Multi-Dimensional Students’ Life Satisfaction Scale, with a higher score indicating higher satisfaction with their family life, friendships, school experience and themselves, where they live, and their overall life \cite{WOS:000179643700001}. Fitness was obtained from a cycle ergometer test which was used to determine the estimated maximal work rate from which VO2max (predicted maximal aerobic power) was estimated. A higher VO2max indicates better aerobic fitness \cite{WOS:A1990DB64800004}.

\section{The Time-Use Optimization Models}
\label{sec:model}

In this section, we first list the notations and descriptions of health outcomes and decision variables in Table \ref{tab:notation}(a). Column \textit{Optimal} lists the definition of optimal value of each objective. Then, we introduce a general model for the one-day time-use optimization problem without considering any specific day structure or health outcome.

\begin{align}
   \textit{obj:} \quad & f(x) = \hat{\beta_0} + \hat{\beta_1} z_1+ \hat{\beta_2} z_2 + \hat{\beta_3} z_3 + \hat{\beta_4} z_1z_1 + \hat{\beta_5} z_1z_2  \nonumber \\
   & + \hat{\beta_6} z_1z_3 + \hat{\beta_7} z_2z_2 + \hat{\beta_8} z_2z_3 + \hat{\beta_9} z_3z_3 \label{eq:objfn} \\
   \textit{s.t.}\quad & z_1 = \sqrt{\frac{3}{4}} \ln \left( \frac{x_1}{\sqrt[3]{x_2 x_3 x_4}} \right),  z_2 = \sqrt{\frac{2}{3}} \ln \left( \frac{x_2}{\sqrt{ x_3 x_4}} \right), z_3 = \sqrt{\frac{1}{2}} \ln \left( \frac{x_3}{x_4} \right) \nonumber\\
    & \sum_{i=1}^4 x_i =1440  \label{con:sum}\\
    &  x_{i}^{l} \leq x_i \leq x_{i}^u \quad \forall i=\{1,\ldots, 4\} \label{con:bounds}
\end{align}

The decision vector of this model can be expressed as $x=\{x_1,x_2,x_3,x_4\}$ which consists of four activity variables (sleep, sedentary time, LPA, MVPA). The objective function \eqref{eq:objfn} shows how to calculate health outcomes based on values of the decision variables and parameters. Where $\beta_0, \beta_1, \ldots, \beta_9$ are unknown regression coefficients to be estimated, they are different in the objective function of each health outcome. Here, those regression coefficients are estimated using the data described in Section \ref{subsec:data}. We list the estimated values $\hat{\beta_i}, i=\{1, \ldots, 9\}$ for different health outcomes in Table \ref{tab:notation} (b) and introduce how to obtain those values in Section \ref{sec:Obtain Beta}. Constraint \eqref{con:sum} forces the sum of decision variables of the problem equal to the total minutes (1440 min) per day. We introduce a closure operation (see Algorithm \ref{alg:closure}) to tackle this constraint and make the working progress of any search algorithm fast to achieve a feasible solution. Upper and lower bounds on each decision variables are enforced by constraint \eqref{con:bounds}, where $x_i^l$ denotes the lower bound of $x_i$ and $x_i^u$ denotes the upper bound of $x_i$. The upper and lower bounds are different according to the day structure considered.

Without loss of generality, we study six different hypothetical day structures. We label these day structures to reflect real-world scenarios: \textit{Studious day (STD)}, \textit{Sporty day (SPD)}, \textit{After-School Job day (ASJD)}, \textit{Sporty Weekend day (SPWD)}, \textit{Studious/screen weekend day (STWD)} and \textit{Working weekend day (WWD)}. The lower and upper bounds of the decision variables are set to suit the day-above-day structures, as advised by an external child behavioural epidemiologist, and by considering the empirical activity durations found in the underlying data (please refer to Table~\ref{tab:bounds}). These replace the 24-hour constraint \eqref{con:bounds} which is present in a general model.

\begin{table}[t]
 \centering
 \caption{Notation and values of parameters}
  \scalebox{0.7}{
\renewcommand{\tabcolsep}{7pt}
\renewcommand{\arraystretch}{1}
    \begin{tabular}{rrrrrrrr}
    \toprule
    \multicolumn{3}{c}{(a) Description of notation}        & \multicolumn{5}{c}{(b) Estimated regression coefficients} \\
    \hline
     \multicolumn{1}{l}{Notation} & \multicolumn{1}{l}{Description } & \multicolumn{1}{l}{Optimal}       & Notation & $f_1$ & $f_2$ &$f_3$& $f_4$\\
          \midrule
    $f_1$ & \multicolumn{1}{l}{Body mass index (BMI)} & $\min|f_1|$        & $\beta_0$  & 0.23307 & 2.3508268 & 12395.053 & 68.85903 \\
           $f_2$ & \multicolumn{1}{l}{Cognition (vocab) objective} & $\max f_2$    &  $\beta_1$ & -0.59691 & -0.032037 & 2255.008 & -17.84326 \\
   $f_3$ & \multicolumn{1}{l}{Life satisfaction objective} & $\max f_3$      & $\beta_2$  & 0.05029 & 0.0670568 & -885.351 & -1.77607 \\
           $f_4$ & \multicolumn{1}{l}{Fitness (VO2max) objective} & $\max f_4$       & $\beta_3$  & 0.68497 & -0.003155 & -1264.635 & -11.25996 \\
                &       &             & $\beta_4$  & 0     & 0     & 0     & 3.15694 \\
  $x_1$ & \multicolumn{2}{l}{Minutes of sleeping} &        $\beta_5$  & 0     & 0     & 0     & 13.88458 \\
           $x_2$ & \multicolumn{2}{l}{Minutes of sedentary behaviour}   & $\beta_6$ & 0     & 0     & 0     & -5.12788 \\
    $x_3$ & \multicolumn{2}{l}{Minutes of LPA} &       $\beta_7$ & 0     & 0     & 0     & -6.85649 \\
           $x_4$ & \multicolumn{2}{l}{Minutes of MVPA} &        $\beta_8$  & 0     & 0     & 0     & 2.69689 \\
                 &       &       & $\beta_9$  & 0     & 0     & 0     & 2.52276 \\
          \bottomrule
    \end{tabular}}%
    \label{tab:notation}
\end{table}%

\begin{table}[t]
  \caption{Values of lower bounds and upper bounds}
\label{tab:bounds}
\centering
 \scalebox{0.75}{
\renewcommand{\tabcolsep}{7pt}
\renewcommand{\arraystretch}{1}
    \begin{tabular}{clrrrrrr}
    \toprule  
     &       & \multicolumn{1}{c}{Studious } & \multicolumn{1}{c}{Sporty} & \multicolumn{1}{c}{After-school} & \multicolumn{1}{c}{Sporty } & \multicolumn{1}{c}{Studious/screen} & \multicolumn{1}{c}{Working} \\
         &       & \multicolumn{1}{c}{day} & \multicolumn{1}{c}{day} & \multicolumn{1}{c}{job day} & \multicolumn{1}{c}{weekend day} & \multicolumn{1}{c}{weekend day} & \multicolumn{1}{c}{weekend day} \\
    \midrule
    \multirow{2}[0]{*}{Sleep} & LB    & 360   & 360   & 360   & 420   & 420   & 360 \\
          & UB    & 720   & 720   & 720   & 720   & 720   & 720 \\
    \multirow{2}[0]{*}{Sedentary} & LB    & 690   & 480   & 480   & 210   & 270   & 210 \\
          & UB    & 900   & 900   & 900   & 900   & 900   & 900 \\
    \multirow{2}[0]{*}{LPA} & LB    & 150   & 210   & 220   & 210   & 150   & 390 \\
          & UB    & 480   & 480   & 480   & 480   & 480   & 480 \\
    \multirow{2}[0]{*}{MVPA} & LB    & 1     & 61    & 1     & 61    & 1     & 1 \\
          & UB    & 210   & 210   & 210   & 210   & 210   & 210 \\
          \bottomrule
    \end{tabular}}%
\end{table}%

\subsection{Model parameter estimation}
\label{sec:Obtain Beta}

Estimates of the model parameters ($\hat{\beta}_i, i = 1,\hdots,9$) in Equation (\ref{eq:objfn}) are calculated using least-squares multiple linear regression on the CheckPoint data. It is not possible to use all the untransformed time-use predictors simultaneously in the linear model as they are \textit{linearly dependent} which in turn prohibits the matrix inverse calculation in estimating the parameter estimates. The \textit{isometric log ratio} (\textit{ilr}) transformation is a widely used transformation of the predictors to remove the linear dependence in the predictors~\cite{WOS:000452307300013}.



For each outcome variable, $f_1,f_2,f_3,f_4$, the Box-Cox transformation is applied after removing predictor effects for variance stabilisation, and improvement in the normality of the residuals \cite{boxcox1964}. Quadratic terms of the time-use \textit{ilr} predictors are considered for each outcome model which correspond to the model terms associated with the parameters $\beta_4, \hdots, \beta_9$ in Equation (\ref{eq:objfn}). If the quadratic terms do not significantly improve the model fit statistically at the $\alpha=0.05$ level (ANOVA $F$-test), the model parameters $\beta_4, \hdots, \beta_9$ are set to 0 (i.e., only linear \textit{ilr} terms remain). For more information about fitting quadratic compositional terms in linear regression, we refer to Chapter 5 of \cite{van2013analyzing}.


The full fit of the linear model also includes covariates of age, sex and puberty status and their associated coefficients. The sample average covariates are then used (age=12, female/male=1:1 and puberty status="Midpubertal"). The estimated effects of these covariates, and the intercept term of the model, are included as the $\beta_0$ term in Equation (\ref{eq:objfn}). The objective functions therefore become the prediction for the theoretical average child in the sample. A sample with missing values in either the outcome or the predictors is removed in each model fit as data are reasonably assumed to be missing at random \cite{saha2005asymptotic}. Diagnostic plots of each model are observed to ensure the model assumptions are reasonable. All analysis is performed in \textsf{R} version 4.0.3 \cite{r403}.

\begin{algorithm}[t]
\caption{Closure Operation}
\label{alg:closure}
\KwIn{Decision vector $\{x_1,x_2,x_3,x_4\}$}
$a = \sum_{i=1}^4 x_{i}$\;
\For {$i=1$ to $4$}{
 $x_{i} = \frac{1440 x_i}{a}$\;
}
\Return the decision variables.
\end{algorithm}

\subsection{One Week Plan}
\label{subsec:oneweek}

We extend the one-day problem to a one-week problem by mixing different day structures, given seven days where each day has four decision variables $x_d =\{x_{d1}, x_{d2}, x_{d3},x_{d4}\}$. Different mixtures shown in Table \ref{tab:oneweekplan} were used to make the one-week plans more realistic. The number listed in each column shows how many of each day type are planned for the week. The objective function for a one-week plan is $F(x) =\sum_{d=1}^7 f(x_d)$ which is subject to the constraints of each included day. 

\begin{table}[t]
 \caption{Different mixture of one-week plan}
\label{tab:oneweekplan}
\centering
 \scalebox{0.8}{
\renewcommand{\tabcolsep}{7pt}
\renewcommand{\arraystretch}{1}
    \begin{tabular}{cccccccc}
    \toprule
    \multicolumn{1}{c}{Index} &       & \multicolumn{1}{c}{Studious } & \multicolumn{1}{c}{Sporty} & \multicolumn{1}{c}{After-school} & \multicolumn{1}{c}{Sporty } & \multicolumn{1}{c}{Studious/screen} & \multicolumn{1}{c}{Working} \\
          &       & \multicolumn{1}{c}{day} & \multicolumn{1}{c}{day} & \multicolumn{1}{c}{job day} & \multicolumn{1}{c}{weekend day} & \multicolumn{1}{c}{weekend day} & \multicolumn{1}{c}{weekend day} \\
          \midrule
    1     &       & 3     & 1     & 0     & 1     & 1     & 1 \\
    2     &       & 3     & 0     & 2     & 0     & 1     & 1 \\
    3     &       & 3     & 2     & 0     & 0     & 1     & 1 \\
    4     &       & 2     & 2     & 1     & 0     & 2     & 0 \\
    5     &       & 2     & 2     & 0     & 1     & 0     & 2 \\
    6     &       & 2     & 2     & 1     & 1     & 1     & 0\\
    \bottomrule
    \end{tabular}}%
  \label{tab:oneweek}%
\end{table}%



\subsection{Multi-objectives Problem}
\label{sec:multi}

Now, we introduce a multi-objective model for time-use optimization. A multi-objective model involves finding solutions to optimize the problem defined by at least two conflicting objectives. The multi-objective model of time-use optimization can be defined as follows.
\begin{align}
   \textit{Objs:} \quad & M(x)=[f_1(x), f_2(x), f_3(x), f_4(x)]
    \label{obj:multi}\\
    s.t. \quad &  \sum_{i=1}^4 x_i =1440 
   \label{con:multisum} \\
   & x_{i}^{l} \leq x_i \leq x_{i}^u \quad \forall i=\{1,\ldots, 4\}
   \label{con:multibounds}
\end{align}
where $x$ denotes a solution, $f_i(x) \rightarrow \mathbb{R}$ denotes the $i$th objective function to be optimized. Since there are four single objectives studied in this paper, we investigate all combinatorial objectives as multi-objective problems.

\subsection{Fitness function}
\label{subsec:singlefit}

We investigate the performance of different evolutionary algorithms for single-objective and multi-objective time-use optimization problems. The fitness of a solution $x$ considers all constraints of one-day time-use optimization and one-week time-use optimization $h(x)$ and $H(x)$ separately.
\begin{align}
    h(x) &= (u(x), f(x)) \label{fit:single} \\
    H(x) &= (U(x), F(x)), \label{fit:multi}
\end{align}
where $u(x)= \sum_{i=1}^4 \max\{0, x_i-x_i^u, x_i^l -x_i\}$ and $U(x) = \sum_{d=1}^7 \sum_{i=1}^4 \max\{0, x_{di}-x_{di}^u, x_{di}^l-x_{di}\}$. We optimize $h$ and $H$ with respect to lexicographic order, i.e. $h(x)\geq h(y)$ holds \textit{iff} $u(x)< u(y) \vee (u(x)=u(y) \wedge f(x) \geq f(y))$ for objective $f_2$, $f_3$ and $f_4$, $u(x)< u(y) \vee (u(x)=u(y) \wedge |f(x)| \leq |f(y)|)$ for objective $f_1$. Therefore, for the time-use optimization problem, any infeasible solution that violates the boundary constraints is worse than any feasible solution. Among solutions that meet all constraints, we aim to optimize the objective function. 

\section{Evolutionary Algorithms for the Time-Use Optimisation Problem}
\label{sec:algoithm}

The algorithms that follow are classified into two classes. The first one contains single-objective evolutionary algorithms (Section \ref{subsec:single}), and the second has multi-objective evolutionary algorithms (Section \ref{subsec:multi}). In this section, we only list the algorithms implemented in this study without detailed descriptions. Moreover, when implementing the presented algorithm for solving time-use optimization, Algorithm \ref{alg:closure} is conducted before evaluating a generated solution.

\subsection{Single-objective evolutionary algorithms}
\label{subsec:single}

For the single-objective time-use optimization, we compare three evolutionary algorithms to optimize all health outcomes in different day structures.

\textbf{Differential Evolution (DE)}\cite{das2010differential,storn1997differential} is a well known global search heuristic using a binomial crossover and a mutation operator. We evaluate two mutation operators \textit{DE/rand/1} and \textit{DE/current-to-rand/1} for the single-objective time -use optimization problem. The population size is set to $50$, and other control parameters are $F=0.5$, $C_r =0.5$.

\textbf{Particle Swarm Optimization (PSO)}\cite{DBLP:journals/tec/AlRashidiE09,DBLP:journals/tec/LeeK13}, is a type of swarm intelligence evolutionary algorithm, with population size $50$, $c_1=1$, $c_2=1$. For more understanding the working processes of PSO, we refer to \cite{banks2007review,banks2008review,DBLP:journals/swevo/HousseinGHS21,kennedy1995particle,wang2018particle,zhang2015comprehensive}.

\textbf{Covariance matrix adaptation evolutionary strategy (CMA-ES)}\cite{DBLP:journals/ec/HansenO01,DBLP:books/sp/06/Hansen06} is a self-adaptive evolution strategy that solves non-linear non-convex optimization problems in continuous domains. We implement the CMA-ES using $\lambda = 10$ and $\sigma =0.3$.

\subsection{Multi-objective evolutionary algorithms}
\label{subsec:multi}

For multi-objective time-use optimization, three multi-objective evolutionary algorithms are considered here.

\textbf{Multi-objective evolutionary algorithm based on decomposition (MOEA/D)} is a decomposition based algorithm commonly used to solve multi-objective optimisation problems \cite{DBLP:journals/tec/ZhangL07}. We use the standard version of MOEA/D with the Tchebycheff approach, and population size is set to $100$.

\textbf{Non-dominated sorting genetic algorithm (NSGA-II)} \cite{996017} is a fast non-dominated sorting procedure for ranking solutions in its selection step. It has been shown to be efficient when dealing with two objective optimization problems. We apply the NSGA-II with SBX operator and set the population size to $100$.

\textbf{Strong pareto evolutionary algorithm 2 (SPEA2)} \cite{zitzler2001spea2} is one of the most popular evolutionary multiple objective algorithms for dealing with optimization problems. We apply the SPEA2 with binary tournament selection and population size $100$.

\begin{table}[t]
   \centering
  \scalebox{0.6}{
  \caption{Mean (mean) and standard deviation (std) of 30 runs (print four decimal places). Best mean values are highlighted in \colorbox{Gray}{\textbf{Best mean}} by comparing results one-day single-objective time-use optimization problem}
  \makebox[\linewidth][c]{
    \begin{tabular}{rlrrrrrrrrrrrrrrrrrrrrr}
    \toprule
    \multicolumn{1}{l}{Day Struct} & Health outcomes &       & \multicolumn{3}{c}{DE/rand/1 (1)} &       & \multicolumn{3}{c}{DE/current-to-rand/1 (2)} &       & \multicolumn{3}{c}{PSO (3)} &       & \multicolumn{3}{c}{CMA-ES (4)} &       & \multicolumn{4}{c}{Best Results} \\
          &       &       & mean  & std   & stat  &       & mean  & std   & stat  &       & mean  & std   & stat  &       & mean  & std   & stat  &       & $x_1$    & $x_2$ & $x_3$ & $x_4$ \\
          \midrule
    \multicolumn{1}{l}{Studious Day} & BMI   &       & 1.8343E-09 & 4.04E-09 & 2,4   &       & 4.6657E-05 & 5.21E-05 & 4     &       & \cellcolor[rgb]{ .816,  .808,  .808}\textbf{1.1039E-13} & 5.82E-13 & 2,4   &       & 0.0012 & 4.41E-19 &       &       & 392 & 713 & 150   & 185 \\
          & Congnition &       & 2.5187 & 4.52E-16 & 2,4   &       & 2.5187 & 1.15E-05 & 4     &       & \cellcolor[rgb]{ .816,  .808,  .808}\textbf{2.5187} & 4.52E-16 & 2,4   &       & 2.5155 & 2.26E-15 &       &       & 389   & 900   & 150   & 1 \\
          & Life satisfaction &       & 12445.2233 & 1.85E-12 & 2,4   &       & 12445.1461 & 0.21  & 4     &       & \cellcolor[rgb]{ .816,  .808,  .808}\textbf{12445.2233} & 1.85E-12 & 2,4   &       & 12331.6566 & 1.85E-12 &       &       & 465 & 690   & 150   & 136 \\
          & Fitness &       & 60.4817 & 4.34E-14 & 4     &       & 60.4817 & 6.34E-14 & 4     &       & \cellcolor[rgb]{ .816,  .808,  .808}\textbf{60.4817} & 4.34E-14 & 4     &       & 60.1741 & 2.17E-14 &       &       & 390   & 690   & 150   & 210 \\
    \multicolumn{1}{l}{Sporty Day} & BMI   &       & 2.4089E-08 & 2.97E-08 & 2,4   &       & 4.6941E-05 & 7.06E-05 & 4     &       & \cellcolor[rgb]{ .816,  .808,  .808}\textbf{3.8719E-16} & 1.40E-15 & 1,2,4 &       & 0.0191 & 3.53E-18 &       &       & 597 & 489 & 210 & 144 \\
          & Congnition &       & 2.4423 & 3.62E-04 & 2,4   &       & 2.4418 & 1.36E-15 &       &       & \cellcolor[rgb]{ .816,  .808,  .808}\textbf{2.4426} & 8.72E-05 & 1,2,4 &       & 2.4419 & 9.03E-16 & 2     &       & 360   & 819 & 210   & 61 \\
          & Life satisfaction &       & 13116.2140 & 9.25E-12 & 2,4   &       & 13116.1700 & 0.09  & 4     &       & \cellcolor[rgb]{ .816,  .808,  .808}\textbf{13118.2140} & 9.25E-12 & 1,2,4 &       & 13044.5263 & 0     &       &       & 573 & 480   & 210   & 178 \\
          & Fitness &       & 62.2440 & 2.17E-14 & 2,4   &       & 62.2343 & 8.68E-03 & 4     &       & \cellcolor[rgb]{ .816,  .808,  .808}\textbf{62.2440} & 2.17E-14 & 2,4   &       & 61.0977 & 2.17E-14 &       &       & 360   & 480   & 390   & 210 \\
    \multicolumn{1}{l}{After-school Job Day} & BMI   &       & \cellcolor[rgb]{ .816,  .808,  .808}\textbf{0.3387} & 0     & 4     &       & 0.3388 & 4.84E-04 & 4     &       & \cellcolor[rgb]{ .816,  .808,  .808}\textbf{0.3387} & 0     & 4     &       & 0.4325 & 2.82E-16 &       &       & 420   & 480   & 330   & 210 \\
          & Congnition &       & 2.4942 & 4.14E-04 &       &       & 2.4932 & 4.39E-04 &       &       & 2.4964 & 8.56E-05 & 1,2   &       & \cellcolor[rgb]{ .816,  .808,  .808}\textbf{2.4964} & 9.03E-16 & 1,2   &       & 360   & 790 & 330   & 1 \\
          & Life satisfaction &       & \cellcolor[rgb]{ .816,  .808,  .808}\textbf{12135.2479} & 3.70E-12 & 2,4   &       & 12135.1980 & 0.07  & 4     &       & \cellcolor[rgb]{ .816,  .808,  .808}\textbf{12135.2479} & 3.70E-12 & 2,4   &       & 12026.5782 & 1.85E-12 &       &       & 481 & 480   & 330   & 149 \\
          & Fitness &       & \cellcolor[rgb]{ .816,  .808,  .808}\textbf{62.2440} & 2.17E-14 & 2,4   &       & 62.2392 & 3.58E-03 & 4     &       & \cellcolor[rgb]{ .816,  .808,  .808}\textbf{62.2440} & 2.17E-14 & 2,4   &       & 61.6931 & 4.34E-14 &       &       & 360   & 480   & 390   & 210 \\
    \multicolumn{1}{l}{Sporty Weekend Day} & BMI   &       & 3.4636E-07 & 3.32E-07 & 2,4   &       & 1.0618E-04 & 7.15E-05 & 4     &       & \cellcolor[rgb]{ .816,  .808,  .808}\textbf{9.2353E-14} & 3.28E-13 & 2,4   &       & 0.0202 & 3.53E-18 &       &       & 718 & 279 & 297 & 146 \\
          & Congnition &       & 2.4338 & 6.71E-04 & 2     &       & 2.4327 & 4.21E-04 &       &       & \cellcolor[rgb]{ .816,  .808,  .808}\textbf{2.4356} & 0     & 1,2,  &       & \cellcolor[rgb]{ .816,  .808,  .808}\textbf{2.4356} & 0     & 1,2   &       & 420   & 784 & 210   & 61 \\
          & Life satisfaction &       & 14453.8050 & 6.57  &       &       & \cellcolor[rgb]{ .816,  .808,  .808}\textbf{14459.7788} & 3.70E-12 & 1,3   &       & 14442.1353 & 1.82E+01 &       &       & \cellcolor[rgb]{ .816,  .808,  .808}\textbf{14459.7788} & 3.70E-12 & 1,3   &       & 720   & 240   & 240   & 210 \\
          & Fitness &       & 60.8883 & 3.49E-03 & 3,4   &       & \cellcolor[rgb]{ .816,  .808,  .808}\textbf{60.8928} & 1.07E-05 & 1,3,4 &       & 60.8739 & 5.79E-04 & 4     &       & 55.8222 & 7.23E-15 &       &       & 441 & 558 & 221 & 210 \\
    \multicolumn{1}{l}{Studious Weekend Day} & BMI   &       & 1.5481E-09 & 1.98E-09 & 2,4   &       & 4.2805E-05 & 5.09E-05 & 4     &       & \cellcolor[rgb]{ .816,  .808,  .808}\textbf{5.9270E-19} & 2.24E-18 & 2,4   &       & 0.0012 & 4.41E-19 &       &       & 458 & 690   & 150   & 142 \\
          & Congnition &       & \cellcolor[rgb]{ .816,  .808,  .808}\textbf{2.5187} & 4.52E-16 & 2,4   &       & 2.5187 & 1.54E-05 & 4     &       & \cellcolor[rgb]{ .816,  .808,  .808}\textbf{2.5187} & 4.52E-16 & 2,4   &       & 2.5155 & 2.26E-15 &       &       & 389   & 900   & 150   & 1 \\
          & Life satisfaction &       & \cellcolor[rgb]{ .816,  .808,  .808}\textbf{12445.2233} & 1.85E-12 & 2,4   &       & 12445.1818 & 0.08  & 4     &       & \cellcolor[rgb]{ .816,  .808,  .808}\textbf{12445.2233} & 1.85E-12 & 2,4   &       & 12331.6566 & 1.85E-12 &       &       & 458 & 690   & 150   & 142 \\
          & Fitness &       & \cellcolor[rgb]{ .816,  .808,  .808}\textbf{60.4817} & 4.34E-14 & 4     &       & 60.4817 & 3.80E-14 & 4     &       & \cellcolor[rgb]{ .816,  .808,  .808}\textbf{60.4817} & 4.34E-14 & 4     &       & 60.1741 & 2.17E-14 &       &       & 390   & 690   & 150   & 210 \\
    \multicolumn{1}{l}{Working Weekend Day} & BMI   &       & 0.0589 & 4.14E-17 & 4     &       & 0.0589 & 2.12E-17 & 4     &       & \cellcolor[rgb]{ .816,  .808,  .808}\textbf{0.0589} & 4.53E-17 & 4     &       & 0.1068 & 7.06E-17 &       &       & 630   & 210   & 390   & 210 \\
          & Congnition &       & 2.4876 & 8.97E-04 &       &       & 2.4858 & 1.40E-03 &       &       & 2.4900 & 5.45E-05 & 1,2   &       & \cellcolor[rgb]{ .816,  .808,  .808}\textbf{2.4900} & 9.03E-16 & 1,2   &       & 360   & 753 & 390   & 1 \\
          & Life satisfaction &       & \cellcolor[rgb]{ .816,  .808,  .808}\textbf{13809.0070} & 5.55E-12 & 2,4   &       & 13808.9369 & 0.10  & 4     &       & \cellcolor[rgb]{ .816,  .808,  .808}\textbf{13809.0070} & 5.55E-12 & 2,4   &       & 13794.5028 & 1.85E-12 &       &       & 641 & 210   & 390   & 199 \\
          & Fitness &       & \cellcolor[rgb]{ .816,  .808,  .808}\textbf{62.2804} & 2.17E-14 & 2,4   &       & 62.2804 & 0     & 4     &       & \cellcolor[rgb]{ .816,  .808,  .808}\textbf{62.2804} & 2.17E-14 & 2,4   &       & 55.7913 & 2.17E-14 &       &       & 360   & 454 & 416 & 210 \\
          \bottomrule
          \label{tab:singelresult}%
    \end{tabular}}}%
\end{table}%

\begin{table*}[t]
  \centering
   \scalebox{0.7}{
  \caption{Mean (mean) and standard deviation (std) of 30 runs (print four decimal places). Best mean values are highlighted in \colorbox{Gray}{\textbf{Best mean}} by comparing results one-week single-objective time-use optimization problem}
  \makebox[\linewidth][c]{
    \begin{tabular}{llrrrrrrrrrrrrrrrr}
    \toprule
    Day Struct & Health outcomes &       & \multicolumn{3}{c}{DE/rand/1 (1)} &       & \multicolumn{3}{c}{DE/current-to-rand/1 (2)} &       & \multicolumn{3}{c}{PSO (3)} &       & \multicolumn{3}{c}{CMA-ES (4)} \\
          &       &       & mean  & std   & stat  &       & mean  & std   & stat  &       & mean  & std   & stat  &       & mean  & std   & stat \\
          \midrule
    1     & BMI   &       & 0.0780 & 3.86E-03 & 2,4   &       & 1.3504 & 0.1327 &       &       & \cellcolor[rgb]{ .816,  .808,  .808}\textbf{0.0657} & 0.0171 & 1,2,4 &       & 0.8056 & 0.1326 & 2 \\
          & Cognition &       & \cellcolor[rgb]{ .816,  .808,  .808}\textbf{17.4336} & 7.12E-06 & 2,3,4 &       & 17.4133 & 0.0045 & 3,4   &       & 17.3504 & 0.0184 & 4     &       & 17.2625 & 0.0268 &  \\
          & Life satisfaction &       & \cellcolor[rgb]{ .816,  .808,  .808}\textbf{91084.2750} & 0.3571 & 2,4   &       & 86871.4104 & 565.3453 &       &       & 91065.6882 & 55.3775 & 2,4   &       & 86950.4715 & 475.0648 &  \\
          & Fitness &       & \cellcolor[rgb]{ .816,  .808,  .808}\textbf{427.1767} & 0.0260 & 2,4   &       & 383.2675 & 3.7084 &       &       & 426.1141 & 3.1869 & 2,4   &       & 406.9240 & 3.5495 & 2 \\
    2     & BMI   &       & 0.7480 & 0.0022 & 2,4   &       & 2.3733 & 0.2167 &       &       & \cellcolor[rgb]{ .816,  .808,  .808}\textbf{0.7368} & 0.0016 & 1,2,4 &       & 1.3006 & 0.0251 & 2 \\
          & Cognition &       & \cellcolor[rgb]{ .816,  .808,  .808}\textbf{17.5453} & 6.34E-06 & 2,3,4 &       & 17.5230 & 0.0043 & 3,4   &       & 17.4626 & 0.0196 & 4     &       & 17.3739 & 0.0239 &  \\
          & Life satisfaction &       & \cellcolor[rgb]{ .816,  .808,  .808}\textbf{87860.0591} & 0.3618 & 2,4   &       & 83699.4248 & 609.0356 &       &       & 87857.4099 & 13.1646 & 2,4   &       & 84854.7392 & 48.1036 & 2 \\
          & Fitness &       & \cellcolor[rgb]{ .816,  .808,  .808}\textbf{428.5592} & 0.0260 & 2,3,4 &       & 378.0875 & 4.7320 &       &       & 423.9175 & 7.3229 & 2,4   &       & 419.0847 & 0.0000 & 2 \\
    3     & BMI   &       & \cellcolor[rgb]{ .816,  .808,  .808}\textbf{0.0742} & 0.0036 & 2,4   &       & 1.5243 & 0.1372 &       &       & 0.0802 & 0.0530 & 2,4   &       & 1.0385 & 0.1753 & 2 \\
          & Cognition &       & \cellcolor[rgb]{ .816,  .808,  .808}\textbf{17.4428} & 5.70E-06 & 2,3,4 &       & 17.4222 & 0.0038 & 3,4   &       & 17.3614 & 0.0244 & 4     &       & 17.2726 & 0.0220 &  \\
          & Life satisfaction &       & \cellcolor[rgb]{ .816,  .808,  .808}\textbf{89821.7778} & 0.5145 & 2,3,4 &       & 85591.7374 & 509.1520 &       &       & 89780.8014 & 97.5659 & 2,4   &       & 85694.0353 & 540.8402 & 2 \\
          & Fitness &       & \cellcolor[rgb]{ .816,  .808,  .808}\textbf{428.5469} & 0.0356 & 2,3,4 &       & 385.2941 & 4.2711 &       &       & 425.9137 & 4.6540 & 2,4   &       & 410.7629 & 4.9936 & 2 \\
    4     & BMI   &       & 0.3538 & 0.0032 & 2,4   &       & 1.7570 & 0.1912 &       &       & \cellcolor[rgb]{ .816,  .808,  .808}\textbf{0.3525} & 0.0393 & 2,4   &       & 0.9861 & 0.1783 & 2 \\
          & Cognition &       & \cellcolor[rgb]{ .816,  .808,  .808}\textbf{17.4516} & 5.87E-06 & 2,3,4 &       & 17.4326 & 0.0042 & 3,4   &       & 17.3689 & 0.0199 & 4     &       & 17.3010 & 0.0118 &  \\
          & Life satisfaction &       & \cellcolor[rgb]{ .816,  .808,  .808}\textbf{88148.1600} & 0.2810 & 3,4   &       & \cellcolor[rgb]{ .816,  .808,  .808}\textbf{88148.1600} & 0.2810 & 3,4   &       & 88144.4983 & 12.2982 & 4     &       & 84944.1618 & 533.3457 &  \\
          & Fitness &       & \cellcolor[rgb]{ .816,  .808,  .808}\textbf{428.5187} & 0.0274 & 2,3,4 &       & 387.9432 & 3.4146 &       &       & 426.5169 & 3.4843 & 2,4   &       & 419.7527 & 5.2723 & 2 \\
    5     & BMI   &       & 0.1364 & 0.0041 & 2,4   &       & 1.4189 & 0.1520 & 4     &       & \cellcolor[rgb]{ .816,  .808,  .808}\textbf{0.1190} & 0.0035 & 1,2,4 &       & 1.5034 & 0.1565 &  \\
          & Cognition &       & \cellcolor[rgb]{ .816,  .808,  .808}\textbf{17.3224} & 2.92E-06 & 2,3,4 &       & 17.3016 & 0.0052 & 3,4   &       & 17.2863 & 0.0169 & 4     &       & 17.1519 & 0.0205 &  \\
          & Life satisfaction &       & \cellcolor[rgb]{ .816,  .808,  .808}\textbf{93118.9359} & 0.5341 & 2,3,4 &       & 88787.2339 & 309.6765 & 4     &       & 93081.6400 & 86.8573 & 2,4   &       & 87702.1107 & 661.4579 &  \\
          & Fitness &       & \cellcolor[rgb]{ .816,  .808,  .808}\textbf{430.6356} & 0.0405 & 2,3,4 &       & 392.2353 & 4.3508 &       &       & 429.6756 & 2.2065 & 2,4   &       & 402.4828 & 4.8568 & 2 \\
    6     & BMI   &       & 0.3587 & 0.0036 & 2,4   &       & 1.5370 & 0.1611 &       &       & \cellcolor[rgb]{ .816,  .808,  .808}\textbf{0.3422} & 0.0097 & 1,2,4 &       & 1.0645 & 0.1762 & 2 \\
          & Cognition &       & \cellcolor[rgb]{ .816,  .808,  .808}\textbf{17.3655} & 4.15E-06 & 2,3,4 &       & 17.3466 & 0.0041 & 3,4   &       & 17.3085 & 0.0220 & 4     &       & 17.1923 & 0.0238 &  \\
          & Life satisfaction &       & \cellcolor[rgb]{ .816,  .808,  .808}\textbf{90081.1467} & 0.6989 & 2,3,4 &       & 86438.6857 & 477.3889 & 4     &       & 90075.1480 & 20.6442 & 2,4   &       & 86225.2387 & 628.1738 &  \\
          & Fitness &       & \cellcolor[rgb]{ .816,  .808,  .808}\textbf{428.8659} & 0.0321 & 2,3,4 &       & 390.9420 & 4.6767 &       &       & 426.5522 & 4.5375 & 2,4   &       & 413.3007 & 4.2733 & 2 \\
          \bottomrule
            \label{tab:singelweekresult}%
    \end{tabular}}}%
\end{table*}%

\begin{table*}[t]
 \centering
  \scalebox{0.65}{
  \caption{Multi-objective optimization hypervolume statistics}
  \makebox[\linewidth][c]{
    \begin{tabular}{llrrrrrrrrrrrrrrrrr}
    \toprule
    Combine of Health Outcomes &       & \multicolumn{5}{c}{MOEA/D (1)}        &       & \multicolumn{5}{c}{NSGA-II (2)}       &       & \multicolumn{5}{c}{SPEA2 (3)} \\
          &       & best  & worst & median & std   & stat  &       & best  & worst & median & std   & stat  &       & best  & worst & median & std   & stat \\
          \midrule
    BMI \& Cognition &       & 0.9895 & 0.9894 & 0.9895 & 7.87E-06 &       &       & 0.9898 & 0.9897 & 0.9898 & 8.67E-06 & 1     &       & \cellcolor[rgb]{ .816,  .808,  .808}\textbf{0.9898} & \cellcolor[rgb]{ .816,  .808,  .808}\textbf{0.9898} & \cellcolor[rgb]{ .816,  .808,  .808}\textbf{0.9898} & 9.13E-06 & 1,2 \\
    BMI \& Life satisfaction &       & 0.9747 & 0.9382 & 0.9451 & 7.34E-03 &       &       & 0.9985 & 0.9984 & 0.9985 & 1.98E-05 & 1     &       & \cellcolor[rgb]{ .816,  .808,  .808}\textbf{0.9988} & \cellcolor[rgb]{ .816,  .808,  .808}\textbf{0.9986} & \cellcolor[rgb]{ .816,  .808,  .808}\textbf{0.9987} & 2.58E-05 & 1,2 \\
    BMI \& Fitness &       & 0.9841 & 0.9837 & 0.9839 & 1.04E-04 &       &       & 0.9841 & 0.9780 & 0.9840 & 1.46E-03 & 1     &       & \cellcolor[rgb]{ .816,  .808,  .808}\textbf{0.9841} & \cellcolor[rgb]{ .816,  .808,  .808}\textbf{0.9841} & \cellcolor[rgb]{ .816,  .808,  .808}\textbf{0.9841} & 6.00E-06 & 1,2 \\
    Cognition \& Life satisfaction &       & 0.9794 & 0.9780 & 0.9788 & 4.10E-04 &       &       & 0.9975 & 0.9967 & 0.9969 & 1.59E-04 & 1     &       & \cellcolor[rgb]{ .816,  .808,  .808}\textbf{0.9978} & \cellcolor[rgb]{ .816,  .808,  .808}\textbf{0.9970} & \cellcolor[rgb]{ .816,  .808,  .808}\textbf{0.9972} & 2.08E-04 & 1,2 \\
    Cognition \& Fitness &       & 0.9959 & 0.9956 & 0.9958 & 6.48E-05 &       &       & 0.9976 & 0.9891 & 0.9952 & 2.48E-03 &       &       & \cellcolor[rgb]{ .816,  .808,  .808}\textbf{0.9977} & \cellcolor[rgb]{ .816,  .808,  .808}\textbf{0.9976} & \cellcolor[rgb]{ .816,  .808,  .808}\textbf{0.9976} & 1.85E-05 & 1,2 \\
    Life satisfaction \& Fitness &       & 0.9961 & 0.9770 & 0.9926 & 7.98E-03 &       &       & 0.9974 & 0.9774 & 0.9970 & 7.07E-03 & 1     &       & \cellcolor[rgb]{ .816,  .808,  .808}\textbf{0.9976} & \cellcolor[rgb]{ .816,  .808,  .808}\textbf{0.9971} & \cellcolor[rgb]{ .816,  .808,  .808}\textbf{0.9973} & 1.53E-04 & 1,2 \\
    BMI \& Cognition \& Life satisfaction &       & 0.9745 & 0.9712 & 0.9714 & 7.58E-04 &       &       & \cellcolor[rgb]{ .816,  .808,  .808}\textbf{0.9874} & \cellcolor[rgb]{ .816,  .808,  .808}\textbf{0.9866} & \cellcolor[rgb]{ .816,  .808,  .808}\textbf{0.9870} & 2.27E-04 & 1,3   &       & 0.9869 & 0.9671 & 0.9816 & 5.16E-03 & 1 \\
    BMI \& Cognition \& Fitness &       & 0.9708 & 0.9690 & 0.9701 & 4.81E-04 &       &       & \cellcolor[rgb]{ .816,  .808,  .808}\textbf{0.9751} & \cellcolor[rgb]{ .816,  .808,  .808}\textbf{0.9724} & \cellcolor[rgb]{ .816,  .808,  .808}\textbf{0.9738} & 7.03E-04 & 1,3   &       & 0.9750 & 0.9396 & 0.9725 & 9.90E-03 &  \\
    Cognition \& Life satisfaction \& Fitness &       & 0.9759 & 0.9572 & 0.9726 & 4.43E-03 &       &       & \cellcolor[rgb]{ .816,  .808,  .808}\textbf{0.9925} & \cellcolor[rgb]{ .816,  .808,  .808}\textbf{0.9780} & \cellcolor[rgb]{ .816,  .808,  .808}\textbf{0.9864} & 4.12E-03 & 1,3   &       & 0.9769 & 0.9607 & 0.9735 & 3.31E-03 &  \\
    BMI \& Cognition \& Life satisfaction \& Fitness &       & 0.9613 & 0.9556 & 0.9569 & 1.33E-03 &       &       & \cellcolor[rgb]{ .816,  .808,  .808}\textbf{0.9706} & \cellcolor[rgb]{ .816,  .808,  .808}\textbf{0.9653} & \cellcolor[rgb]{ .816,  .808,  .808}\textbf{0.9683} & 1.27E-03 & 1,3   &       & 0.9677 & 0.9463 & 0.9601 & 5.07E-03 &  \\
    \bottomrule
     \label{tab:bioresults}%
    \end{tabular}}}%
\end{table*}%

\section{Experiments}
\label{sec:exp}

This section shows detailed optimization results comparing the different evolutionary algorithms. Firstly, to evaluate the performance of the single-objective algorithms we investigate one-day instances of six different day structures with boundary constraints (Table \ref{tab:bounds}) against four single objectives. Secondly, we evaluate the performance of the multi-objective algorithms on six different mixtures of one-week instances (Table \ref{tab:oneweek}), taking \textit{Sporty day} as an exemplar with all the combinations of objectives for bio-objective optimization.

For each optimization algorithm with the configurations above, we execute 30 runs and report the statistic results using the Kruskal-Wallis test with $95\%$ confidence intervals and follow-up with Bonferroni adjustments to account for multiple comparisons \cite{Corder2014}. All experiments are performed using Jmetal of version 5.11, which is based on the description included in \cite{DBLP:conf/gecco/NebroDV15}, and carried out on a MacBook Pro with an M1 chip.

\subsection{Results of Single-objective Time-Use Optimization}

\begin{figure*}[t]
\centering
\subfloat[Median HV Run of optimizing \textit{BMI} and \textit{ Cognition}]{\includegraphics[width=0.45\linewidth]{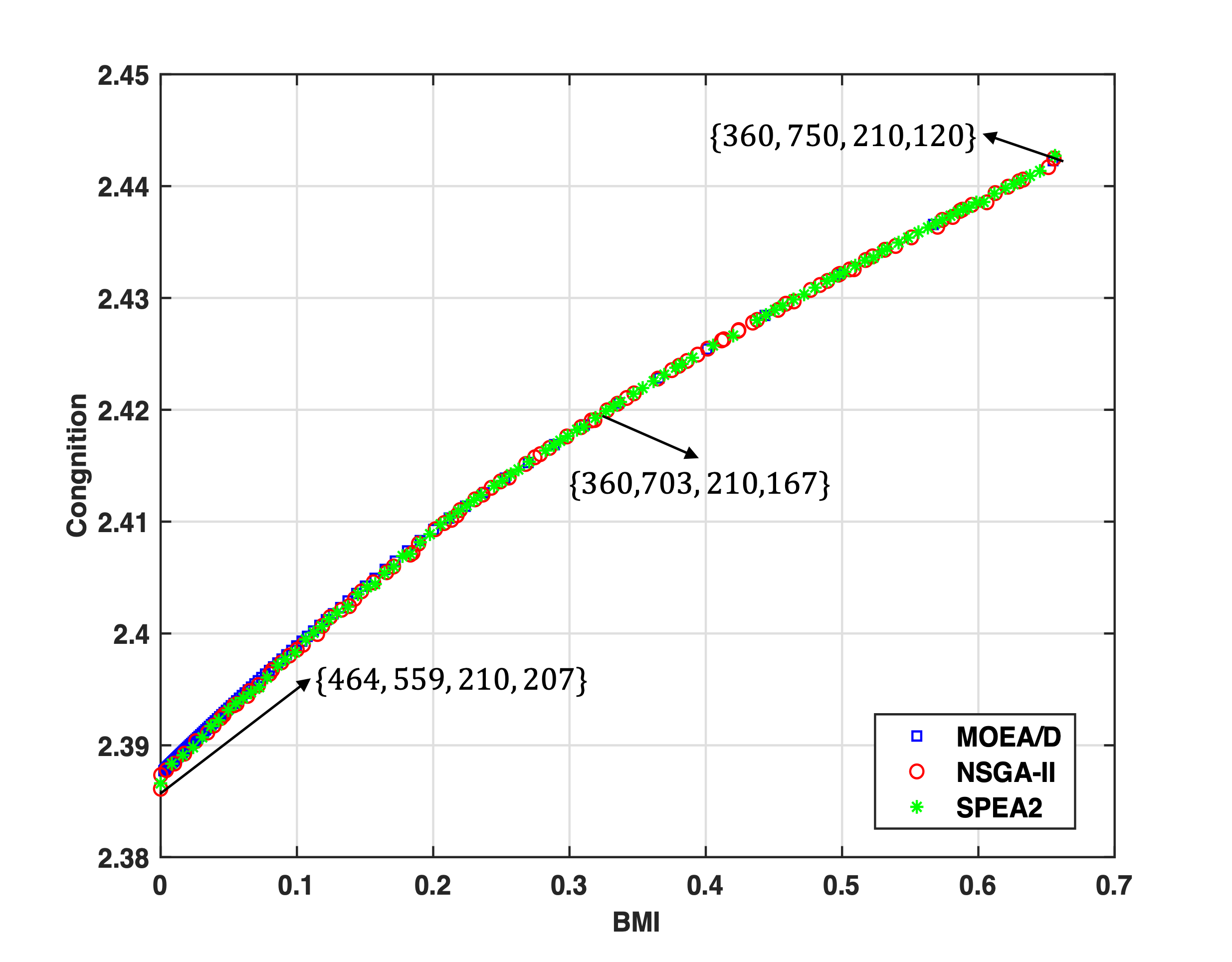}}
\hfill
\subfloat[Median HV Run of optimizing \textit{Life satisfaction} and \textit{Fitness}]{\includegraphics[width=0.45\linewidth]{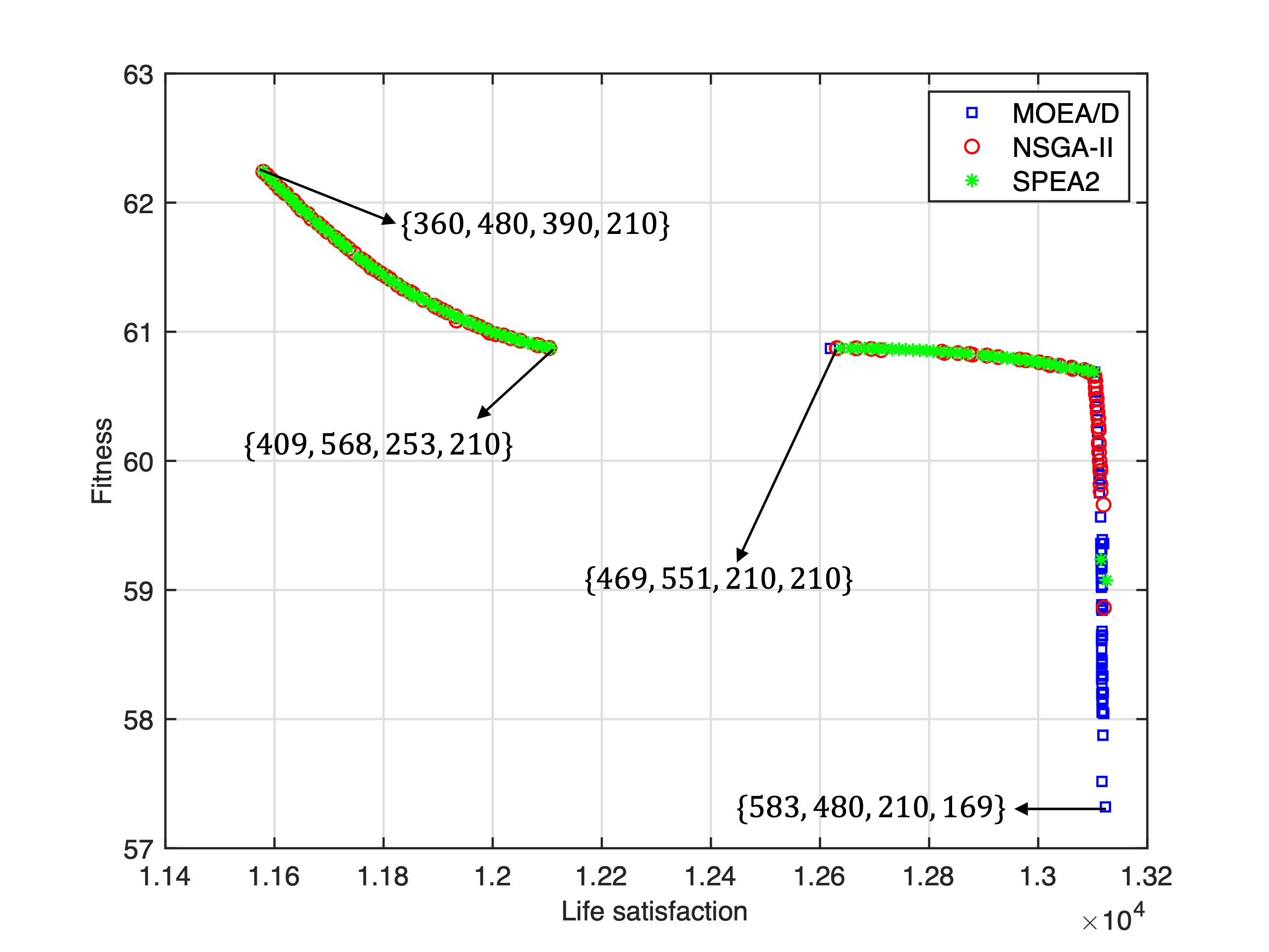}}
\\
\subfloat[Median HV Run of optimizing \textit{BMI}, \textit{Cognition} and \textit{Life satisfaction}]{\includegraphics[width=0.45\linewidth]{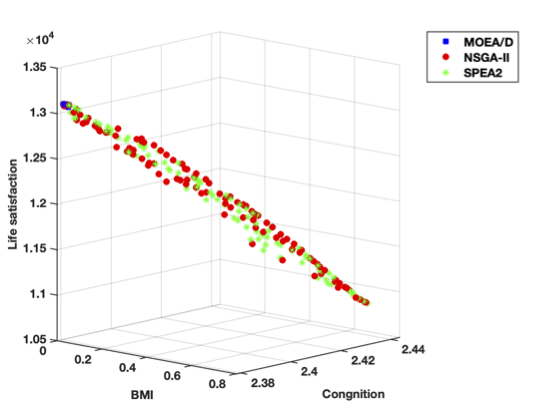}}
\hfill
\subfloat[Median HV Run of optimizing \textit{Cognition}, \textit{Life satisfaction} and \textit{Fitness}]{\includegraphics[width=0.45\linewidth]{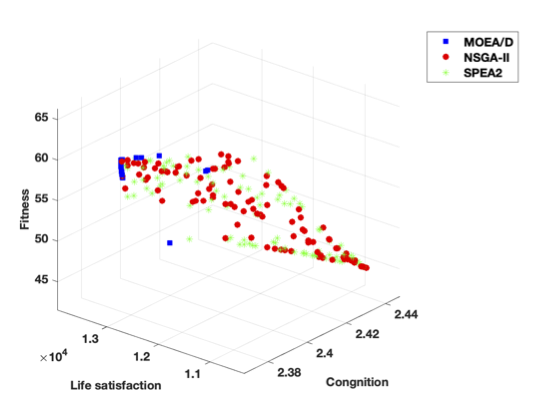}}
\caption{Results obtained for multi-objective model of sporty day}
\label{fig:HVplots}
\end{figure*}

Table \ref{tab:singelresult} and Table \ref{tab:singelweekresult} list the results obtained of one-day instances and one-week instances separately. We provide the results from 30 independent runs with $25,000$ generation for all instances. The \textit{mean} denotes the average objective value of the 30 runs and \textit{std} denotes standard deviation. Since we aim to minimize the absolute value of BMI, the results listed in the \textit{BMI} rows are absolute values. The best solutions are bold and shadowed in each row. We also report the decision variables (rounded to minutes) of the optimal solution for each health outcome of each day structure in Table \ref{tab:singelresult}.

Column \textit{stat} lists the results of statistical comparisons between the algorithms. If two algorithms can be compared significantly, then the index of algorithms that list in each column is significantly worse than the current algorithm. For example, the first row in Table \ref{tab:singelresult} shows that PSO and DE/rand/1 are significantly better than DE/current-to-rand/1 and CMA-ES when optimizing the BMI of Studious day, and DE/current-to-rand/1 is significantly better than CMA-ES. However, there is no significant difference between the performance of DE/rand/1 and PSO. As can be seen from the table of one-day instances, the results obtained by the PSO are better than other algorithms in nearly all cases. DE algorithm with \textit{DE/rand/1} operator is the second best algorithm, outperforming the DE algorithm with \textit{DE/current-to-rand/1} operator and CMA-ES in many instances, while CMA-ES shows an advantage when aiming to optimize Cognition for many day structures. Moreover, as observed in the \textit{std} columns, the standard deviation of 30 runs of all the evaluated algorithms in most instances is close to zero. Therefore, we can argue that for the single-objective  optimization, the results obtained by the investigated algorithms, especially the DE/rand/1 and PSO, are close to optimal.

Table \ref{tab:singelweekresult} presents the summary statistic for the results of one-week single-objective instances. A closer inspection of the table shows that DE/rand/1 outperforms the other algorithms in most instances, and PSO outperforms the last two algorithms. Therefore, these results suggest that for solving the single-objective  optimization problem, DE algorithm with \textit{DE/rand/1} operator and PSO both perform well. PSO is preferred for one-day instances, and the DE algorithm with \textit{DE/rand/1} operator is preferred for solving one-week instances.

\subsection{Results of Multi-objective Time-Use Optimization}

To compare the difference between evolutionary multi-objective optimization algorithms, we analyze the experimental results of two, three and four objectives, respectively. For performance evaluation, we use hypervolume \cite{DBLP:journals/tec/ZitzlerT99,DBLP:journals/tcs/AugerBBZ12} as the metric. The hypervolume statistics are provided in Table \ref{tab:bioresults}. The best hypervolume is highlighted and bold for each combination of objectives in each row. It can be seen from the \textit{ stat} results in the table that SPEA significantly outperforms the other algorithms for two-objective optimization instances, and NSGA-II outperforms the other two algorithms for three- and four-objective optimization instances.

The bio-objective results obtained in a median hypervolume run for each algorithm are plotted in Figure \ref{fig:HVplots}. Fig. \ref{fig:HVplots} (a) shows that the trade-off fronts of optimizing the first two objectives achieved by SPEA2 are more generally distributed in the Pareto front than MOEA/D and NSGA-II. Similarly, Fig. \ref{fig:HVplots} (b) indicates that the trade-off solutions obtained by MOEA/D and NSGA-II are clustered in a small area of the solution space. Moreover, for three-objective optimization (Fig. \ref{fig:HVplots} (c) and (d)), NSGA-II and SPEA2 generate better Pareto solutions in comparison with MOEA/D. On Fig. \ref{fig:HVplots} (a) and (b), selected optimized  solutions are shown to reflect optimal daily activity durations if one individual outcome is preferred above another (near to the respective axes) or if the outcomes are equally preferred (near the mid-point of the Pareto front).

\section{Conclusion}
\label{sec:con}
The way children spend their time on sleep, sedentary behaviour and physical activity (LPA and MVPA) affects their health and well-being. The main goal of the current study is to implement evolutionary algorithms on daily  allocations to optimize children's health outcomes. Based on a real-world data set, we introduce single- and multi-objective  optimization models and design fitness functions of one-day and one-week problems. Our experimental results show that when tackling the single-objective problem, DE algorithm with \textit{DE/rand/1} and PSO outperforms other proposed algorithms on both one-day instances and one-week instances. Moreover, the SPEA2 has a higher hypervolume than NSGA-II and MOEA/D in two-objective optimization instances for the multi-objective problem. In comparison, NSGA-II has a higher hypervolume than the other algorithms in three and four objectives instances. Overall, this study strengthens the idea that evolutionary algorithms can be used to enhance our understanding of how children can allocate their daily time to optimize their health and well-being. Parents are concerned about their children's sleep, screen time and physical activity, and they want evidence-based guidance on how much time should be spent in these behaviours. However, it is unlikely to be feasible to expect families to follow strict daily time allocation schedules. The evidence generated from the application of optimization algorithms may be better understood as general advice, and primarily serve to inform public health guidelines for children's time-use behaviours. Population-level surveillance of guideline compliance can help inform public health policy, track secular trends overtime and to evaluate the effectiveness of public health interventions.

\section{Acknowledgements}
This work has been supported by NHMRC Ideas grant 1186123, by ARC grant FT200100536, and by the South Australian Government through the Research Consortium "Unlocking Complex Resources through Lean Processing". Dorothea Dumuid is supported by NHMRC Fellowship 1162166 and by the Centre of Research Excellence in Driving Global Investment in Adolescent Health funded by NHMRC 1171981. The CheckPoint study was supported by the NHMRC [1041352; 1109355]; the National Heart Foundation of Australia [100660]; The Royal Children’s Hospital Foundation [2014-241]; the Murdoch Children’s Research Institute (MCRI); The University of Melbourne; the Financial Markets Foundation for Children [2014-055, 2016-310]; and the Australian Department of Social Services (DSS). Research at the MCRI is supported by the Victorian Government's Operational Infrastructure Support Program. The funders played no role in the study design, data collection and analysis, decision to publish, or preparation of the manuscript.

\bibliographystyle{abbrv}  
\bibliography{main}

\end{document}